\documentclass{article}

\usepackage{microtype}
\usepackage{graphicx}
\usepackage{subfigure}
\usepackage{booktabs} % for professional tables

\usepackage{amsmath}
\usepackage{tikz}
\usetikzlibrary{positioning}
\usetikzlibrary{calc}

\usepackage{hyperref}

\usepackage[accepted]{icml2020}

\icmltitlerunning{Deep Claim: Payer Response Prediction from Claims Data with Deep Learning}

\begin{document}

\twocolumn[
\icmltitle{Deep Claim: Payer Response Prediction from Claims Data with Deep Learning}



\icmlsetsymbol{equal}{*}

\begin{icmlauthorlist}
\icmlauthor{Byung-Hak Kim}{ah}
\icmlauthor{Seshadri Sridharan}{ah}
\icmlauthor{Andy Atwal}{ah}
\icmlauthor{Varun Ganapathi}{ah}
\end{icmlauthorlist}

\icmlaffiliation{ah}{Alpha Health, South San Francisco, CA, USA}

\icmlcorrespondingauthor{Byung-Hak Kim}{hak.kim@alphahealth.com}

\icmlkeywords{Machine Learning, ICML}

\vskip 0.3in
]



\printAffiliationsAndNotice{}  % leave blank if no need to mention equal contribution

\begin{abstract}
Each year, almost 10\% of claims are denied by payers (i.e., health insurance plans). With the cost to recover these denials and underpayments, predicting payer response (likelihood of payment) from claims data with a high degree of accuracy and precision is anticipated to improve healthcare staffs' performance productivity and drive better patient financial experience and satisfaction in the revenue cycle~\citep{Barkholz17}. However, constructing advanced predictive analytics models has been considered challenging in the last twenty years. That said, we propose a (low-level) context-dependent compact representation of patients' historical claim records by effectively learning complicated dependencies in the (high-level) claim inputs. Built on this new latent representation, we demonstrate that a deep learning-based framework, Deep Claim, can accurately predict various responses from multiple payers using 2,905,026 de-identified claims data from two US health systems. Deep Claim's improvements over carefully chosen baselines in predicting claim denials are most pronounced as 22.21\% relative recall gain (at 95\% precision) on Health System A, which implies Deep Claim can find 22.21\% more denials than the best baseline system.
% as it traditionally requires manual feature extraction and rules discovery from various sources of noisy data. 
% In other words, this implies with the level of precision (if Deep Claim flags 100 claims, 95 of them or more are denials), Deep Claim can find 22.21\% more denials than the best baseline system. 
% We also demonstrate that Deep Claim can be used to flag suspicious fields of the claim to provide actionable insights for healthcare staff to revisit prior to the submission.
\end{abstract}

\section{Introduction}
\label{sec:introduction}
A recent study found that nearly a quarter of annual healthcare spending is wasteful, with the most extensive source being administrative expenses, totaling \$266 billion per year~\citep{Shrank19}. 
Hundreds of thousands of medical insurance claims are submitted by hospitals each day, and payers initially deny about 5-11\% of hospital claims. For the average hospital in the US, this statistic means about \$5 million in payments are at risk each year. Moreover, while 63\% of denials can be recovered, it approaches \$120 per claim in administrative costs to recoup the monies owed~\citep{Medicare17, Barkholz17, Carroll20}. 

In the last twenty years, claims are prepared and submitted mainly by humans. There does not exist any predictive system to review claims before submission in the revenue cycle. Consequently, hospitals and health systems are eager to adopt automation to every step of the revenue cycle. By this automation, health organizations can decrease transaction time, increase savings, and redirect staff to focus effort toward more critical tasks that provide the most significant patient care value~\citep{Pecci19, Pecci20}. 

The first step toward this end is an automated machine learning (ML) system that enables healthcare providers to reliably predict which claims are to be denied and which forecasts a payer's response date to a claim even before the claim has been submitted. These predictions have the aim of guiding revenue cycle staff, focusing attention on high-value denials and those that have a strong likelihood of being overturned, and even prompting staff to correct claims before submission to increase first-pass payment rates. It is easy to see how a portion of the \$266 billion in the annual waste can be potentially reduced by a well-developed automated ML system.

Unlike electronic health records (EHR), claims data are by its nature temporally bounded and are produced primarily for the administration of payment for health services delivered by healthcare providers and facilities. 
% It is not designed to convey information about what happened in the past. 
Additionally, since all health care providers want to be paid for their services, nearly every interaction a patient has with any medical system leads to the generation of a billing claim, resulting in an abundant and standardized patient information source. %Claims data broadly capture information from all doctors and providers caring for a patient, though it is usually limited to diagnosis and procedures.

In this paper, we propose Deep Claim, a deep learning-based framework that predicts payers' responses to claims. As far as we know, this is the first deep learning system that successfully addresses the problem, and our work makes the three contributions below: 
\begin{itemize}
    \item First, Deep Claim compresses patient-level information and models complicated \textbf{clinical contextual interrelations} in the (high-level) claims data targeted to payers' response prediction tasks (through multi-task deep learning). Unlike prior works that have focused on the most predictive features per payer, Deep Claim instead exploits different payers' raw claims data directly by leveraging a gating mechanism and bilinear models without depending on expert domain knowledge and significant data preprocessing. 
    \item Second, based on data from two US health systems with a general patient population and diverse payer mixes, we demonstrate the effectiveness of the Deep Claim framework in \textbf{real deployment scenarios}. 
    \item Third, in addition to the payer response predictions, Deep Claim can identify the claim's most questionable fields that should be reviewed. This \textbf{prediction interpretability} is critical in real use in providing the health administrators some level of explanations with an automated machine predictions.  
\end{itemize}

\section{Related Work}
\label{sec:related works}
\textbf{Traditional Scrubber-based Approach:} Billed amounts are typically determined based on contracts between providers and payers. The contracts define the amounts to be paid for services provided, and customized scrubbing software tools (or sometimes humans) update each payer's adjudication rules to verify charge information before a claim is sent to the payer and checking for any potential errors~\citep{Umair09}.

\textbf{Data-driven Rule-based Models:} Another line of study is to discover attributional rules from data that can be used to screen claims by detecting irregularities before the claim is submitted to payers~\citep{Anand08, Kumar10, Bradley10, Ghani11, Wojtusiak11, Saripalli17}. The most recent work~\citet{Saripalli17} used engineered features based on Claim Adjustment Reason Codes (CARC)~\citep{carc:2019} and applied several classification methods, including a neural network (NN) model. However, the NN model had shown inferior performance as compared to non-NN approaches (e.g., Decision Trees). This is mainly because the models rely on feature engineering to handle sparse and large input dimensions, which appears to limit one to develop better NN models.
 
\textbf{Patient Representations Learning:} More recent research focused on developing patient representations using embedding ideas to handle sparsity of the raw patient data~\citep{Miotto16, Choi16, Rajkomar18, Choi18, Zhong19, Bai19}. Many studies, including ~\citet{Miotto16,  Rajkomar18, Choi18}, primarily focused on EHR data, so it is not straightforward to use them to the claims data. Notably, as shown in ~\citet{Miotto16}, an unsupervised based approach often looses predictive power in prediction tasks. Alternatively, ~\citet{Choi18} performed the joint learning of the hidden structure of EHR data and a supervised prediction task and showed better performance (than unsupervised embedding and external domain knowledge). However, this performance boost was attained by exploiting EHR’s encounter structure, which is entirely unavailable in the claims data. 

\section{Deep Claim}
\label{sec:Deep Claim}
\begin{figure}[ht]
    \centering
    \vspace{-0.35cm}
        \includegraphics[width = 0.38\textwidth]{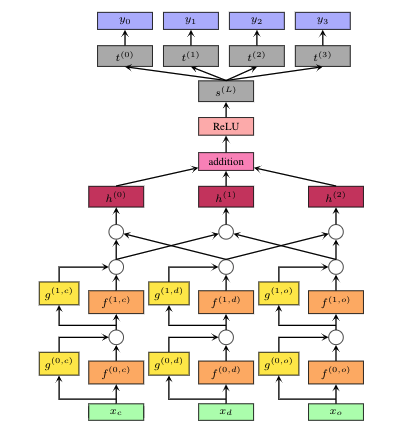}
    \vspace{-0.40cm}
    \caption{Architecture of a Deep Claim for the payer response prediction as described in Section \ref{sec:Deep Claim}.}
    \label{fig:DeepClaimArch}
\end{figure}

We propose Deep Claim as a neural network-based system to predict whether, when, and how much a payer will pay for each claim. Deep Claim takes the claims data composed of demographic information, diagnoses, treatments, and billed amounts as an input. Given that, Deep Claim predicts the first response date, denial probability, denial reason codes with probability, and questionable fields in the claim. In this section, we describe the Deep Claim model in detail, which the complete architecture illustrated in Figure~\ref{fig:DeepClaimArch}.

% \begin{figure}[ht]
%     \vspace{-0.25cm}
%     \resizebox{0.4\textwidth}{!}{\input{arch5.tikz}}
%     \vspace{-0.25cm}
%     \caption{Architecture of a Deep Claim for the payer response prediction as described in Section \ref{sec:Deep Claim}.}
%     \label{fig:DeepClaimArch}
% \end{figure}
\vspace{-0.25cm}

\subsection{Claims Input Representation}
\label{ssec:claims representation}
The claim vector we create from the raw claim is composed of a huge number of variables (i.e., features) - subscriber gender, an individual relationship code, a payer state, the duration of the corresponding service, the subscriber's age, the patient's age, a payer identifier, the total charges, the services date, and transmission of the claim date. The claim vector also includes an indication of procedures performed and diagnoses received. The value of each feature is assigned a single unique token for singular elements or sub-context vectors of tokens for procedures and diagnoses (that can have multiple values). 

We tokenize procedures and diagnoses and map them to a sub-context vector of tokens. Less frequent tokens are mapped to an out-of-vocabulary (OOV) token (for example, procedure token appears less than 500 times in the dataset). We also normalize numeric values. The date is mapped to tokens in years, months, and days. The charge amount in dollars is quantized to thousands, hundreds, tens, and ones. The patient's age is discretized in years.

After defining the features, we categorize them into three \emph{contextual} categories: procedure, diagnosis, and other features regarding the claim, such as demographic patient information. Procedures and diagnosis token vectors can be expressed as a normalized count vector (e.g., relative frequency) \(\mathbf{x_{c}}\) and \(\mathbf{x_{d}}\) with a length of the possible procedure and diagnosis tokens respectively. All the other single unique feature tokens can be comprised as \(\mathbf{x_{o}}\), which is a binary vector of a length of the total number of single unique tokens. One can piece all of them together to convert a single claim to a concatenated vector \(\mathbf{x}\) as \((\mathbf{x_{c}},\mathbf{x_{d}},\mathbf{x_{o}})\). Typically, this vector \(\mathbf{x}\) can have a length in the thousands and be the extremely sparse vector.

\subsection{Claims Embedding Network}
\label{ssec:claims embedding}
Unlike natural language sentences, the extremely sparse vector \(\mathbf{x}\) is an unordered collection of medical events and aggregations of diverse code types that encapsulates various aspects of complicate dependencies. So it is not straightforward to apply off-the-shelf NLP embedding techniques for compressing this sparse vector into a fixed-sized latent vector \(\mathbf{h}\) (94 in our experiments). Instead, we leverage gating mechanism, which is essential for recurrent neural networks~\citep{Hochreiter97,Cho14} and bilinear models~\citep{Tenenbaum00,Kim17} that provide richer representations than linear models. To be specific, we propose the following novel methods to learn effective embedding representation mappings \(H:\mathbf{x}\rightarrow{}\mathbf{h}\) of each claim by activating the gate over each context sub-vector to extract inter-component dependencies within each category and combining them further to learn intra-dependencies among the context sub-vectors by taking the pairwise inner product in the latent low-dimensional space. 

First, we convert each sub-category vector to lower-dimensional context vectors \(\mathbf{f^{(0,i)}}\) simply as \(\sigma(\mathbf{W^{(0,i)}_{f}}\mathbf{x_{i}}+\mathbf{b^{(0,i)}_{f}})\) where \(\mathbf{W_{f}}\) is the low-dimensional embedding matrix and \(\sigma\) is a ReLU~\citep{Nair10} function for \(\mathbf{i=\{c,d,o\}}\). 
Then, the context vectors \(\mathbf{c^{(0,i)}}\) modulated by the gates is represented as \(\mathbf{f^{(0,i)}} \odot g^{(0,i)}(\mathbf{W^{(0,i)}_{g}}\mathbf{x_{i}}+\mathbf{b^{(0,i)}_{g}})\) where \(g\) is the Softmax function and \(\odot\) denotes element-wise multiplication. These gate activation values over each sub-vector can be viewed as dynamic importance scores of the (high-level) input feature that enables learnable feature selection and simultaneous dimensionality reduction while handling sparsity in each sub-vector. To further increase the hierarchy of gated layers like a probabilistic decision tree, we add one more set of gated networks for \(\mathbf{c^{(1,c)}}\) in the (low-level) latent space. Next, to capture inter-dependencies among the multiple clinical context categories (we show here for three categories for brevity), we include pairwise multiplicative layers and add all pairwise representations to project back to the joint latent-context space and apply ReLU to obtain \(\mathbf{h}\) depicted as \(\sigma(h^{(0)}(\mathbf{c^{(1,c)}} \odot \mathbf{c^{(1,d)}}) + h^{(1)}(\mathbf{c^{(1,o)}} \odot \mathbf{c^{(1,c)}}) + h^{(2)}(\mathbf{c^{(1,c)}} \odot \mathbf{c^{(1,d)}}))\) where a function \(h^{(i)}(\mathbf{x}) \) defined as \(BN^{(i)}(\mathbf{W^{(i)}}\mathbf{x})\). Note that including BN~\citep{Ioffe15} in a later stage appears effective as adding another layer of recombination and addition encourages to learn agreed representations among the three contexts.

\subsection{Multi-Task Learning Network}
\label{ssec:multi-task learning}
Claim embedding \(\mathbf{h}\) can be trained for any prediction tasks. Since we are interested in the payer response predictions, we select outcomes using the associated payment (remittance) response information (e.g., 835) sent by the payers to the providers. We focus on the CARC codes which communicate why a claim or service line was paid differently than it was billed. We first compile a set of most frequently occurring CARC codes that constitute denials observed in the dataset\footnote{We use 14 different reason codes for denial of the claim on Health System A and 19 codes on Health System B in our experiments. These sets of codes were used to label the dataset, and additional manual labeling is not required.}. 

Given this list, we induce three outcomes - claim denial variable \(y_{0}\) which is 1 only iff claim or service level CARC codes found in the corresponding remittance fall in the set and normalized reason code count vectors \(\mathbf{y_{1}}\) (in claim level) and \(\mathbf{y_{2}}\) (in service level). Note \(y_{0}\) is a single variable and \(\mathbf{y_{1}}\) and \(\mathbf{y_{2}}\) are normalized counts in frequency. Additionally, we estimate the first response date variable \(y_{3}\) which is a day interval between remittance date and the corresponding claim submission date. Concatenating these together gives the target outcome vector \(\mathbf{y}\) as \((y_{0}, \mathbf{y_{1}}, \mathbf{y_{2}}, y_{3})\) which denotes a probability that the claim will be denied under a set of possible denial reason codes in how many days.

Provided \(\mathbf{x}\) and \(\mathbf{y}\) pairs, Deep Claim applies \(\mathbf{s^{(l)}}\) transformation defined as \(\sigma(\mathbf{W^{(l)}_{s}}\mathbf{h}+\mathbf{b^{(l)}_{s}})\) \(L\) times, then each task can have an (optional) individual tower of network \(\mathbf{t^{(j)}}\) on top of \(\mathbf{s^{(L)}}\) as \(\sigma(\mathbf{W^{(j)}_{t}}\mathbf{s^{(L)}}+\mathbf{b^{(j)}_{t}})\). After, we set up the following optimization problem to have a claim embedding captured satisfactorily by sharing the bottom layers while keeping task-specific top layers as:
\vspace{-0.2cm}
\begin{align*}
  \arg\min_{(\mathbf{h},H,\mathbf{W})} &\null \mathcal{L}_{BCE}(y_{0},\sigma(\mathbf{W_{0}}\mathbf{t^{(0)}}+\mathbf{b_{0}})) \\ 
  +\,&\null \lambda_{0}\,\mathcal{L}_{CCE}(\mathbf{y_{1}},\text{softmax}(\mathbf{W_{1}}\mathbf{t^{(1)}}+\mathbf{b_{1}})) \\
  +\,&\null \lambda_{1}\,\mathcal{L}_{CCE}(\mathbf{y_{2}},\text{softmax}(\mathbf{W_{2}}\mathbf{t^{(2)}}+\mathbf{b_{2}})) \\
  +\,&\null \lambda_{2}\,\mathcal{L}_{L_{1}}(y_{3},\mathbf{W_{3}}\mathbf{t^{(3)}}+\mathbf{b_{3}}), 
\end{align*}
% \vspace{-0.25cm}
where \(\mathcal{L}_{BCE}\) is the binary cross-entropy loss and \(\mathcal{L}_{CCE}\) is the categorical cross-entropy loss. First two constraints are for possible denial reason codes classifications (in both service and claim levels) that we call denial reason codes constraints and the last constraint is the response time constraint for predicting payer response date. These added three constraints act as barrier functions to guide the convergence to a better embedding (see Section~\ref{sec:Experimental Results} for more details). 

\subsection{Predictions Interpretability}
\label{ssec:predictions interpretability}
In addition to generating a prediction regarding the outcome of submitted claims, Deep Claim can predict the questionable field of the claim that should be reviewed. Our goal is to determine which aspects of the claim would most strongly cause the denial decision to be flipped. To that end, we compute the normalized gradient magnitude of the prediction score (between 0 and 1) for the input feature dimension using a single back-propagation pass\footnote{This idea was inspired by a saliency map-based visualization technique initially developed for the Computer Vision~\citep{Simonyan14}.}. We define the suspiciousness score of a claim field to be the sensitivity of the outcome to the value of that particular field. This suspiciousness score represents the contribution of an input feature on the denial prediction of a corresponding claim. With these scores, Deep Claim can identify input features that contribute most to a denial prediction of a given claim. Or the system could flag input features with suspiciousness scores above a predetermined threshold (e.g., 0.8) such that users may review and modify claim data.  

\section{Experimental Results} 
\label{sec:Experimental Results}
To highlight the merits of each building block of Deep Claim in the spirit of ablation study, we evaluate two Deep Claim models, two variants, and two baseline models\footnote{Due to the space limit, see the Supplementary Material for details on the experimental results: Section 1 for evaluation design, Section 2 for table results, and Section 3 and 4 for the figures.}. Table 1 and Table 2 clearly show that the performance gap between Deep Claim models becomes substantial as we take out each core block. Notably, on Health System A, the best Deep Claim model (0.8045 PR-AUC) demonstrates significant improvement over the best baseline (0.7813 PR-AUC), resulting in 22.21\% relative recall gain. Comparing the first three rows of Table 3 suggests how much the added constraints in multi-task learning effectively help improve recall accuracy. 

Additionally, we evaluate the payer response date prediction. We use Mean Absolute Error (MAE) to measure average distance from true response time. Table 4 shows the best Deep Claim model provides 23.9\% relative MAE reduction in the response date prediction as compared to a simple baseline of an average on Health System A, which is highly desirable (the average baseline provides MAE of 6.421 day for Health System A and 5.232 day for Health System B). 

Lastly, as discussed in Section~\ref{ssec:predictions interpretability}, Figure 4 presents an example plot of suspiciousness score vs. claim field of the claim sequence of a claim. We use a Deep Claim model on Health System A for this visualization. By reading off points showing higher suspiciousness scores, one can recognize what would be the strong causes (e.g., missing or incorrect information) for potential claim denial. The plot also clearly shows three segments following \(\mathbf{x_{c}}\), \(\mathbf{x_{d}}\), and \(\mathbf{x_{o}}\) category order which indicate how much each sub-vector of \(\mathbf{x}\) would contribute to the prediction. We observed that those structured details and focused view of top five suspicious fields in Figure 4 resulted partly from having the denial reason code constraints active by setting \((\lambda_{0}=1.0, \lambda_{1}=1.0\)). 

\section{Discussion} 
In this paper, we have developed a neural network-based Deep Claim framework for payer response prediction, which had received little attention before. Our brand-new framework does not need any hand-selection of features (including payer-specific data harmonization) deemed predictive by a domain expert. It is capable of learning (low-level) \textit{clinical} context-dependent claim embeddings from (high-level) raw claim data (through gated layers followed by multiplicative interactions in multi-task learning) and further providing a list of questionable fields of the claim. With this Deep Claim system, the opportunity for a healthcare organization to free up revenue cycle staff time for more critical projects and to drive cost savings is significant. Although in potential product deployment, we need to provide means to prevent potential negative consequences of using the system with fraudulent intent to facilitate fraudulent activities. We hope that our results will catalyze new developments of deep learning technologies to automate every step of the revenue cycle process in healthcare administration. 

\textbf{Limitations:} We have not covered how much the reported prediction performance of Deep Claim brings a positive impact on the healthcare systems. In fact, the computed suspicion scores can be visualized (e.g., on the CMS-1500 claim form) to inform what the suspicious fields of the claim are to revisit. As a next step, we intend to evaluate how much this visualized insight collectively with the likelihood of denial would help administration teams not to waste time working on the wrong claims. On top of making more financially efficient processing of claims, we also would like to assess the full benefits of this work in the healthcare loop towards reducing patient's financial burdens and improving the quality of care. We want to leave studies and discussions in this direction as to future work. 

\section*{Acknowledgements}
We thank Woosang Lim for helpful conversations on a low-rank bilinear pooling layer.

% If a paper is accepted, the final camera-ready version can (and
% probably should) include acknowledgements. In this case, please
% place such acknowledgements in an unnumbered section at the
% end of the paper. Typically, this will include thanks to reviewers
% who gave useful comments, to colleagues who contributed to the ideas,
% and to funding agencies and corporate sponsors that provided financial


\bibliography{icml}
\bibliographystyle{icml2020}

%%%%%%%%%%%%%%%%%%%%%%%%%%%%%%%%%%%%%%%%%%%%%%%%%%%%%%%%%%%%%%%%%%%%%%%%%%%%%%%
% \end{document}

\twocolumn[
\icmltitle{Deep Claim: Supplementary Materials}
\vskip 0.3in
]
\setcounter{section}{0}
\section{Evaluation Design} 
\subsection{Data Preparation}
\label{ssec:data preparation}
The dataset was constructed using the historical 837 and 835 claims data provided by two very different (e.g., size, payer mixes) US health systems. We extracted only the necessary information from 837 and 835 separately and pieced them together based on a unique patient control number to create 1,267,527 records from Health System A and 1,637,499 records from Health System B. It should be noted that no personally identifiable information is included in this data and we deliberately chose not to include demographic information analysis since claims data is usually deidentified.

The claim vector \(\mathbf{x}\) is produced following the steps illustrated in Section 3.1. The resulting length is greater than 3000\footnote{\((\text{len}(\mathbf{x_{c}}), \text{len}(\mathbf{x_{d}}), \text{len}(\mathbf{x_{o}}))\)=(921,1582,1205) for Health System A and \((\text{len}(\mathbf{x_{c}}), \text{len}(\mathbf{x_{d}}), \text{len}(\mathbf{x_{o}}))\)=(805,1597,646) for Health System B.}. We map procedure tokens appearing less than 500 times, in both Health System A and B, to OOV tokens. Less frequent diagnosis tokens (i.e., appears less than 400 times in Health System A, and less than 800 times in Health System B) are also encoded to OOV tokens. As described in Section 3.3, a denial event is defined as any claim or service level reason codes fall into the most frequent (and meaningful) denial reason codes set. Health systems disclosed typical claim denial rate range but did not disclose actual reason codes used for their own sake. Thereby, we had to compile a set of relevant reason codes to label each claim and verified the list with them. Intending to simulate typical health system scenarios, we use the code set size of 14 to induce claim denial rate of 14.31\% on Health System A and 20 for the rate of 12.17\% on Health System B.

\begin{table*}[]
    \caption{Recall value at 95\% precision level of each model for claim denial prediction on Health System A and B. Values in parenthesis indicate standard deviations from 3-split time series CV. The bold value shows the highest value for each column condition. Note that the best Deep Claim model has shown 22.21\% relative recall gain over the best baseline on Health System A.}\vspace{0.5cm}
    \label{table: REC95-HSAB}
    \centering
    \begin{tabular}{lccc}
    \toprule[\heavyrulewidth]
    \textbf{Model} & \textbf{Health System A} & \textbf{Health System B} \\
    \midrule
    DeepClaim1(L=2,w/ towers) & 0.4746 (0.0269) & \textbf{0.3898} (0.0572) \\ 
    DeepClaim2(L=2,w/o towers) & \textbf{0.4748} (0.0470) & 0.3841 (0.0624) \\ 
    DeepClaim2(L=0,w/o towers)-multipliers & 0.4499 (0.0448) & 0.3555 (0.0584) \\ 
    DeepClaim2(L=0,w/o towers)-multipliers-gates & 0.4305 (0.0543) & 0.3440 (0.0823) \\ 
    Baseline, NN model & 0.3494 (0.0722) & 0.3483 (0.0288) \\ 
    Baseline, Random Forest model & 0.3885 (0.0156) & 0.3051 (0.0551) \\ 
    \bottomrule[\heavyrulewidth]
    \end{tabular}
\end{table*}

\begin{table*}[]
    \caption{PR-AUC of each model for claim denial prediction on Health System A and B. Values in parenthesis indicate standard deviations from 3-split time-series CV. The bold value shows the highest value for each column condition. Note that we report PR-AUC values, not as a performance metric, but to showcase the quantitative differences between models.}  \vspace{0.5cm}
    \label{table: PRAUC-HSAB}
    \centering
    \begin{tabular}{lccc}
    \toprule[\heavyrulewidth]
    \textbf{Model} & \textbf{Health System A} & \textbf{Health System B} \\
    \midrule
    DeepClaim1(L=2,w/ towers) & 0.7985 (0.0478) & 0.7841 (0.0194) \\ 
    DeepClaim2(L=2,w/o towers) & \textbf{0.8045} (0.0419) & \textbf{0.7844} (0.0179) \\ 
    DeepClaim2(L=0,w/o towers)-multipliers & 0.7977 (0.0431) & 0.7788 (0.0225) \\ 
    DeepClaim2(L=0,w/o towers)-multipliers-gates & 0.7999 (0.0452) & 0.7801 (0.0203) \\ 
    Baseline, NN model & 0.7822 (0.0440) & 0.7639 (0.0210) \\ 
    Baseline, Random Forest model & 0.7813 (0.0443) & 0.7723 (0.0249) \\ 
    \bottomrule[\heavyrulewidth]
    \end{tabular}
\end{table*}

\begin{table*}[]
    \caption{Recall value at 95\% precision level of Deep Claim model provided different sets of \((\lambda_{0},\lambda_{1},\lambda_{2})\) for claim denial prediction on Health System A and B. Values in parenthesis indicate standard deviations from 3-split time-series CV. Note that the first three rows imply how much the added constraints in multi-task learning effectively help improve recall accuracy upon the best baseline.} \vspace{0.5cm}
    \label{table: REC95-LAMDA-HSAB}
    \centering
    \begin{tabular}{lccc}
    \toprule[\heavyrulewidth]
    \textbf{Model} & \textbf{Health System A} & \textbf{Health System B} \\
    \midrule
    DeepClaim1, \((\lambda_{0},\lambda_{1},\lambda_{2})=(1.0,1.0,0.01)\) & 0.4746 (0.0269) & 0.3898 (0.0572) \\ 
    DeepClaim1, \((\lambda_{0},\lambda_{1},\lambda_{2})=(1.0,1.0,0.0)\) & 0.4650 (0.0025) & 0.3694 (0.0396) \\ 
    DeepClaim1, \((\lambda_{0},\lambda_{1},\lambda_{2})=(0.0,0.0,0.0)\) & 0.4323 (0.0722) & 0.3554 (0.0715) \\ 
    Baseline, Random Forest model & 0.3885 (0.0156) & 0.3051 (0.0551) \\ 
    \bottomrule[\heavyrulewidth]
    \end{tabular}
\end{table*}

\begin{table*}[]
    \caption{MAE of each model for the payer response date prediction on Health System A and B. Values in parenthesis indicate standard deviations from 3-split time-series CV. The bold value shows the lowest value for each column condition.}  \vspace{0.5cm}
    \label{table: MAE-HSAB}
    \centering
    \begin{tabular}{lccc}
    \toprule[\heavyrulewidth]
    \textbf{Model} & \textbf{Health System A} & \textbf{Health System B} \\
    \midrule
    DeepClaim1(L=2,w/ towers) & 4.9146 (0.8138) & 3.6359 (0.0429) \\ 
    DeepClaim2(L=2,w/o towers) & \textbf{4.8835} (0.7786) & 3.6413 (0.0748) \\ 
    DeepClaim2(L=0,w/o towers)-multipliers & 5.0773 (1.1330) & \textbf{3.6191} (0.0908) \\ 
    DeepClaim2(L=0,w/o towers)-multipliers-gates & 4.9450 (0.9837) & 3.6209 (0.0630) \\ 
    Baseline, NN model & 5.0205 (0.7343) & 3.6254 (0.1082) \\ 
    % Baseline, average & 6.421 & 5.232\\ 
    \bottomrule[\heavyrulewidth]
    \end{tabular}
\end{table*}

\subsection{Training}
\label{ssec:training details}
For the Deep Claim model training, we use the ADAM \citep{Kingma15} optimizer with \(lr = 0.001\) and \((\beta_{1},\beta_{2})=(0.9,0.999)\). We set \((\lambda_{0},\lambda_{1},\lambda_{2})=(1.0,1.0,0.01)\). 
Across the models, we use the same weight matrix \(\mathbf{W}\) of 96 dimension to make a fair comparison. Further hyper-parameter optimization could be done for the optimal accuracy of Deep Claim model. 

We carefully chose a Random Forest model and a single layer NN model as two baselines. They were shown to be effective in previous works~\citep{Saripalli17,Choi18}, so considered as a representative approach of a non-NN and NN baseline model. We did not consider unsupervised baseline, which shown in ~\citep{Miotto16} to lose the accuracy in some prediction tasks. For fair comparisons, the non-NN baseline performance was optimized extensively by grid searching hyperparameters (e.g., number of trees in the forest). 

\subsection{Evaluation Metrics}
\label{ssec:eval metricss}
To validate the benefits of the Deep Claim model, we evaluate for the payer response prediction under a real production scenario and metric as illustrated below.

\textbf{Time Series Cross-Validation (CV):} As the submission date characterizes the claims data, it is imperative to evaluate Deep Claim model not using traditional cross-validation. To accurately simulate the real world prediction environment, in which we train in the present and predict the future, we split data temporally so that the test set data comes chronologically after the training set. In that end, we use a variation of \(k\)-fold, which returns first past \(k\) folds as train set and the future \((k+1)\)-th fold as a test set. Note that unlike standard CV methods, successive training sets are supersets of those that come before them, and the error on each split is averaged to compute a robust estimate of model error. 

\textbf{Precision-Recall Curve (PRC):} Since the true binary target label of claim denial is highly imbalanced; we use the PRC as an alternative to Receiver Operating Characteristic (ROC) for evaluating the quality of the predictions. This choice is because PRC can show performance differences between balanced and imbalanced datasets, and it is more useful in revealing the early-retrieval area (i.e., low false-positive rate area in the ROC plot) performance~\citep{Saito15}. In a real deployment scenario, the system returning accurate results (high precision), as well as returning a majority of all positive results (high recall), is most desirable, so we measure recall value at a very high precision level of 95\% (if the system flags 100 claims, 95 of them or more are denials) as a key performance metric.

% \vspace{-0.25cm}
\section{Tables of Results} 
\label{sec:Tables of Results}
Here we present the full quantitative table results.

\section{Figures of Architectures Used in the Experiments} 
\label{sec:Figures}
\begin{figure}[ht]
    \centering 
    \usetikzlibrary{shapes.geometric, shapes.multipart, arrows, calc, positioning}

\tikzset{
input/.style = {draw, fill=green!30, rectangle, minimum width=1.25cm, minimum height=0.10cm},
block/.style = {draw, fill=orange!60, rectangle, minimum width=1.25cm, minimum height=0.10cm},
gate/.style= {draw, fill=yellow!100, rectangle, minimum width=0.5cm, minimum height=0.10cm, node distance=1cm},
mul/.style= {draw, fill=white, circle, minimum width=0.1cm, minimum height=0.1cm, node distance=1cm},
blk_h/.style = {draw, fill=purple!80, rectangle, minimum width=1.25cm, minimum height=0.10cm},
blk_r/.style = {draw, fill=red!30, rectangle, minimum width=1.25cm, minimum height=0.10cm},
blk_i/.style = {draw, fill=magenta!60, rectangle, minimum width=1.25cm, minimum height=0.10cm},
blk_t/.style = {draw, fill=black!30, rectangle, minimum width=1.25cm, minimum height=0.10cm},
output/.style = {draw, fill=blue!30, rectangle, minimum width=1.25cm, minimum height=0.10cm}
}

\tikzstyle{arrow} = [thick,->,>=stealth]

\begin{tikzpicture}[node distance=0.5cm]
    \tikzstyle{every node}=[font=\scriptsize]
    \node [input, name=rinput] (rinput) {};

    \node (xc)      [input]{$x_{c}$};    
    \node (f00)     [block, above of=xc, yshift=0.4cm]{$f^{(0,c)}$};
    \node (gate00)  [gate, left of=f00, xshift=-0.3cm, yshift=0.2cm]{$g^{(0,c)}$};
    \node (mul00)   [mul, above of=f00, yshift=-0.2cm]{};
    \node (f10)     [block, above of=mul00, yshift=0.3cm]{$f^{(1,c)}$};
    \node (gate10)  [gate, left of=f10, xshift=-0.3cm, yshift=0.2cm]{$g^{(1,c)}$};
    \node (mul10)   [mul, above of=f10, yshift=-0.2cm]{};
    % \node (mul0)    [mul, above of=mul10, yshift=-0.2cm]{};
    \node (h0)      [blk_h, above of=mul10, yshift=0.3cm]{$h^{(0)}$};

    \node (xd)      [input, right of=xc, xshift=2.0cm]{$x_{d}$};
    \node (f01)     [block, above of=xd, yshift=0.4cm]{$f^{(0,d)}$};
    \node (gate01)  [gate, left of=f01, xshift=-0.3cm, yshift=0.2cm]{$g^{(0,d)}$};
    \node (mul01)   [mul, above of=f01, yshift=-0.2cm]{};
    \node (f11)     [block, above of=mul01, yshift=0.3cm]{$f^{(1,d)}$};
    \node (gate11)  [gate, left of=f11, xshift=-0.3cm, yshift=0.2cm]{$g^{(1,d)}$};
    \node (mul11)   [mul, above of=f11, yshift=-0.2cm]{};
    \node (h1)      [blk_h, above of=mul11, yshift=0.3cm]{$h^{(1)}$};

    \node (xo)      [input, right of=xd, xshift=2.0cm]{$x_{o}$};
    \node (f02)     [block, above of=xo, yshift=0.4cm]{$f^{(0,o)}$};
    \node (gate02)  [gate, left of=f02, xshift=-0.3cm, yshift=0.2cm]{$g^{(0,o)}$};
    \node (mul02)   [mul, above of=f02, yshift=-0.2cm]{};
    \node (f12)     [block, above of=mul02, yshift=0.3cm]{$f^{(1,o)}$};
    \node (gate12)  [gate, left of=f12, xshift=-0.3cm, yshift=0.2cm]{$g^{(1,o)}$};
    \node (mul12)   [mul, above of=f12, yshift=-0.2cm]{};
    \node (h2)      [blk_h, above of=mul12, yshift=0.3cm]{$h^{(2)}$};

    \node (add)     [blk_i, above of=h1, yshift=0.3cm]{addition};
    \node (relu)    [blk_r, above of=add, yshift=0.3cm]{ReLU};
    \node (y0)    [output, above of=relu, xshift=-2.3cm, yshift=0.3cm]{${y}_{0}$};
    \node (y1)    [output, right of=y0, xshift=1.0cm]{${y}_{1}$};
    \node (y2)    [output, right of=y1, xshift=1.0cm]{${y}_{2}$};
    \node (y3)    [output, right of=y2, xshift=1.0cm]{${y}_{3}$};
    
    \draw [arrow] (xc) -- (f00);
    \draw [arrow] (f00) -- (mul00);
    \draw [arrow] ([yshift=0.1cm]xc.north) -| (gate00);
    \draw [arrow] (gate00) |- (mul00);
    \draw [arrow] (mul00) -- (f10);
    \draw [arrow] (f10) -- (mul10);
    \draw [arrow] ([yshift=0.1cm]mul00.north) -| (gate10);
    \draw [arrow] (gate10) |- (mul10);
    
    \draw [arrow] (xd) -- (f01);
    \draw [arrow] (f01) -- (mul01);
    \draw [arrow] ([yshift=0.1cm]xd.north) -| (gate01);
    \draw [arrow] (gate01) |- (mul01);
    \draw [arrow] (mul01) -- (f11);
    \draw [arrow] (f11) -- (mul11);
    \draw [arrow] ([yshift=0.1cm]mul01.north) -| (gate11);
    \draw [arrow] (gate11) |- (mul11);
    
    \draw [arrow] (xo) -- (f02);
    \draw [arrow] (f02) -- (mul02);
    \draw [arrow] ([yshift=0.1cm]xo.north) -| (gate02);
    \draw [arrow] (gate02) |- (mul02);
    \draw [arrow] (mul02) -- (f12);
    \draw [arrow] (f12) -- (mul12);
    \draw [arrow] ([yshift=0.1cm]mul02.north) -| (gate12);
    \draw [arrow] (gate12) |- (mul12);
    
    \draw [arrow] (mul10) -- (h0);
    \draw [arrow] (mul11) -- (h1);
    \draw [arrow] (mul12) -- (h2);
    
    \draw [arrow] (h0.north) -- (add);
    \draw [arrow] (h1.north) -- (add);
    \draw [arrow] (h2.north) -- (add);
    
    \draw [arrow] (add) -- (relu);
    \draw [arrow] (relu.north) -- (y0.south);
    \draw [arrow] (relu.north) -- (y1.south);
    \draw [arrow] (relu.north) -- (y2.south);
    \draw [arrow] (relu.north) -- (y3.south);

\end{tikzpicture}
    \caption{\textbf{DeepClaim2(L=0,w/o towers)-multipliers} architecture used in the experiments. Color-coding is the same as in Figure 1. Note that this architecture is to manifest the impact of the pairwise multiplicative layers in the Deep Claim model.}
    \label{fig:Arch8}
\end{figure}
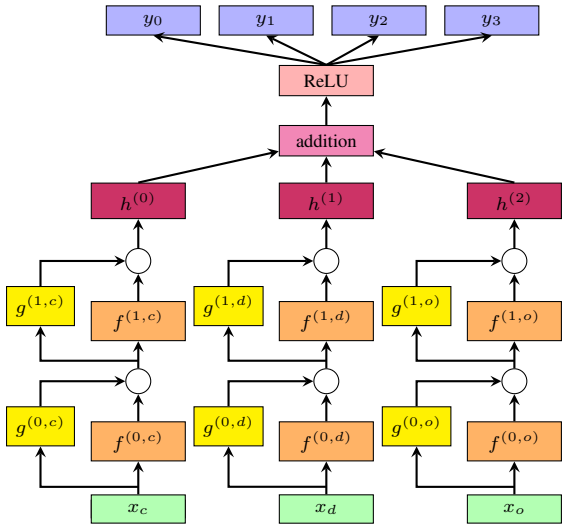

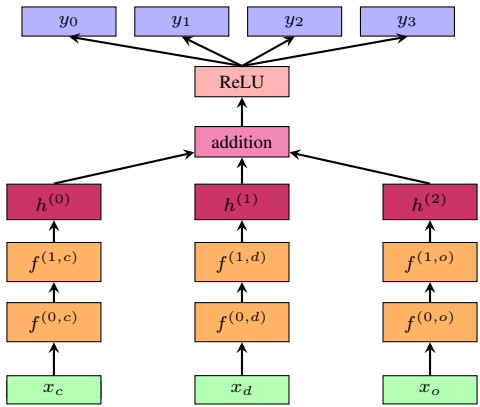
\begin{figure}[ht]
    \centering 
    \usetikzlibrary{shapes.geometric, shapes.multipart, arrows, calc, positioning}

\tikzset{
input/.style = {draw, fill=green!30, rectangle, minimum width=1.25cm, minimum height=0.10cm},
block/.style = {draw, fill=orange!60, rectangle, minimum width=1.25cm, minimum height=0.10cm},
gate/.style= {draw, fill=yellow!100, rectangle, minimum width=0.5cm, minimum height=0.10cm, node distance=1cm},
mul/.style= {draw, fill=white, circle, minimum width=0.1cm, minimum height=0.1cm, node distance=1cm},
blk_h/.style = {draw, fill=purple!80, rectangle, minimum width=1.25cm, minimum height=0.10cm},
blk_r/.style = {draw, fill=red!30, rectangle, minimum width=1.25cm, minimum height=0.10cm},
blk_i/.style = {draw, fill=magenta!60, rectangle, minimum width=1.25cm, minimum height=0.10cm},
blk_t/.style = {draw, fill=black!30, rectangle, minimum width=1.25cm, minimum height=0.10cm},
output/.style = {draw, fill=blue!30, rectangle, minimum width=1.25cm, minimum height=0.10cm}
}

\tikzstyle{arrow} = [thick,->,>=stealth]

\begin{tikzpicture}[node distance=0.5cm]
    \tikzstyle{every node}=[font=\scriptsize]
    \node [input, name=rinput] (rinput) {};

    \node (xc)      [input]{$x_{c}$};    
    \node (f00)     [block, above of=xc, yshift=0.4cm]{$f^{(0,c)}$};
    \node (f10)     [block, above of=f00, yshift=0.3cm]{$f^{(1,c)}$};
    \node (h0)      [blk_h, above of=f10, yshift=0.3cm]{$h^{(0)}$};

    \node (xd)      [input, right of=xc, xshift=2.0cm]{$x_{d}$};
    \node (f01)     [block, above of=xd, yshift=0.4cm]{$f^{(0,d)}$};
    \node (f11)     [block, above of=f01, yshift=0.3cm]{$f^{(1,d)}$};
    \node (h1)      [blk_h, above of=f11, yshift=0.3cm]{$h^{(1)}$};

    \node (xo)      [input, right of=xd, xshift=2.0cm]{$x_{o}$};
    \node (f02)     [block, above of=xo, yshift=0.4cm]{$f^{(0,o)}$};
    \node (f12)     [block, above of=f02, yshift=0.3cm]{$f^{(1,o)}$};
    \node (h2)      [blk_h, above of=f12, yshift=0.3cm]{$h^{(2)}$};

    \node (add)     [blk_i, above of=h1, yshift=0.3cm]{addition};
    \node (relu)    [blk_r, above of=add, yshift=0.3cm]{ReLU};
    \node (y0)    [output, above of=relu, xshift=-2.3cm, yshift=0.3cm]{${y}_{0}$};
    \node (y1)    [output, right of=y0, xshift=1.0cm]{${y}_{1}$};
    \node (y2)    [output, right of=y1, xshift=1.0cm]{${y}_{2}$};
    \node (y3)    [output, right of=y2, xshift=1.0cm]{${y}_{3}$};
    
    \draw [arrow] (xc) -- (f00);
    \draw [arrow] (f00) -- (f10);
    \draw [arrow] (f10) -- (h0);

    \draw [arrow] (xd) -- (f01);
    \draw [arrow] (f01) -- (f11);
    \draw [arrow] (f11) -- (h1);

    \draw [arrow] (xo) -- (f02);
    \draw [arrow] (f02) -- (f12);
    \draw [arrow] (f12) -- (h2);

    \draw [arrow] (h0.north) -- (add);
    \draw [arrow] (h1.north) -- (add);
    \draw [arrow] (h2.north) -- (add);
    
    \draw [arrow] (add) -- (relu);
    \draw [arrow] (relu.north) -- (y0.south);
    \draw [arrow] (relu.north) -- (y1.south);
    \draw [arrow] (relu.north) -- (y2.south);
    \draw [arrow] (relu.north) -- (y3.south);

\end{tikzpicture}
    \caption{\textbf{DeepClaim2(L=0,w/o towers)-multipliers-gates} used in the experiments. Color-coding is the same as in Figure 1. Note that this architecture is to demonstrate the impact of the gating mechanisms in the Deep Claim model.}
    \label{fig:Arch9}
\end{figure}

\begin{figure}[ht]
    \centering 
    \usetikzlibrary{shapes.geometric, shapes.multipart, arrows, calc, positioning}

\tikzset{
input/.style = {draw, fill=green!30, rectangle, minimum width=1.25cm, minimum height=0.10cm},
block/.style = {draw, fill=orange!60, rectangle, minimum width=1.25cm, minimum height=0.10cm},
blk_i/.style = {draw, fill=magenta!60, rectangle, minimum width=1.25cm, minimum height=0.10cm},
gate/.style= {draw, fill=yellow!100, rectangle, minimum width=0.5cm, minimum height=0.10cm, node distance=1cm},
output/.style = {draw, fill=blue!30, rectangle, minimum width=1.25cm, minimum height=0.10cm}
}

\tikzstyle{arrow} = [thick,->,>=stealth]

\begin{tikzpicture}[node distance=0.5cm]
    \tikzstyle{every node}=[font=\scriptsize]
    \node [input, name=rinput] (rinput) {};

    \node (xc)    [input]{$x_{c}$};    
    \node (xd)    [input, right of=xc, xshift=2.0cm]{$x_{d}$};
    \node (xo)    [input, right of=xd, xshift=2.0cm]{$x_{o}$};

    \node (concat)   [blk_i, above of=xd, yshift=0.3cm]{concatenation};
    \node (f)     [block, above of=concat, yshift=0.4cm]{$f$};

    \node (y0)    [output, above of=f, xshift=-2.3cm, yshift=0.3cm]{${y}_{0}$};
    \node (y1)    [output, right of=y0, xshift=1.0cm]{${y}_{1}$};
    \node (y2)    [output, right of=y1, xshift=1.0cm]{${y}_{2}$};
    \node (y3)    [output, right of=y2, xshift=1.0cm]{${y}_{3}$};
    
    \draw [arrow] (xc.north) -- (concat);
    \draw [arrow] (xd.north) -- (concat);
    \draw [arrow] (xo.north) -- (concat);
    
    \draw [arrow] (concat) -- (f);
    
    \draw [arrow] (f.north) -- (y0.south);
    \draw [arrow] (f.north) -- (y1.south);
    \draw [arrow] (f.north) -- (y2.south);
    \draw [arrow] (f.north) -- (y3.south);

\end{tikzpicture}
    \caption{\textbf{Baseline, NN model} used in the experiments. Note that this baseline architecture is to show the advantage of Deep Claim over the standard bag of feature-based aggregation approach.}
    \label{fig:Arch0}
\end{figure}
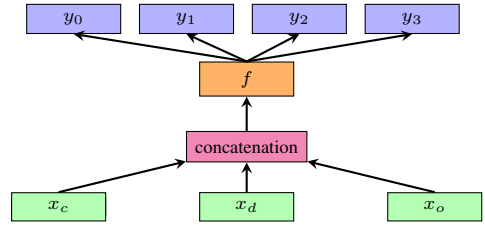

\vspace{4.00cm}
\section{An Example Plot of Suspiciousness Score} 
\label{sec:Suspiciousness Score Figures}
\begin{figure}[ht]
    \centering
    \includegraphics[width = 0.50\textwidth]{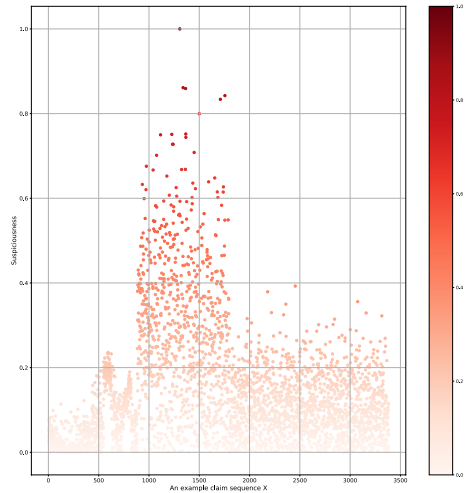}
    \vspace{-0.40cm}
    \caption{A plot of suspiciousness score vs. claim field of the claim sequence for a Health System A's example claim case. Note that a Deep Claim model predicts this example claim to be denied and presents five fields in diagnosis subvector that contribute most (showing higher than 80\% scores) to a denial prediction of a claim.}
    \label{fig:HSAscore}
\end{figure}

\end{document}